# FUZZY AND ENTROPY FACIAL RECOGNITION

Jae Jun Lee[1] and Taeseon Yoon[2]

[1]Department of Natural Science, Hankuk Academy of Foreign Studies, Yong-In, Republic Korea
chjjma@naver.com
[2]Hankuk Academy of Foreign Studies, Yong-In, Republic Korea
tsyoon@hafs.hs.kr

*ABSTRACT*

*This paper suggests an effective method for facial recognition using fuzzy theory and Shannon entropy. Combination of fuzzy theory and Shannon entropy eliminates the complication of other methods. Shannon entropy calculates the ratio of an element between faces, and fuzzy theory calculates the membership of the entropy with 1. More details will be mentioned in Section 3. The learning performance is better than others as it is very simple, and only need two data per learning. By using factors that don't usually change during the life, the method will have a high accuracy.*

*KEYWORDS*

*Fuzzy, Entropy, Facial Recognition, Shannon Entropy, Bell-curve*

## 1. INTRODUCTION

Face recognition from still image is one of the active research areas with numerous law enforcement applications. Previous attempts to develop facial recognition system include Hidden Markov Models based on the extraction of 2D-DCT feature vectors [4], correlation method [5], and eigenface method [6]. In this paper, a new method of facial recognition system combining Fuzzy theory and Shannon entropy will be introduced. Fuzzy Theory is a theory that deals with subjects that are uncertain by describing the position of something for a situation that don't have only one answer, just like the things happening in our real life. The biggest benefit of using fuzzy theory is that we can apply on our real life, because our real life is full of uncertainness. So, fuzzy theory is used on areas that need to reflect our real life such as artificial intelligence, pattern recognition, automatic control, etc. Fuzzy theory has been used a lot for facial recognition system such as fuzzy-based segment-boost method [1], fisherface method with fuzzy membership degree [2], 3D Face Recognition in the Multiple-Contour Line Area Using Fuzzy Integral [3]. Shannon entropy, which was defined by Shannon, is a method to calculate unpredictability of elements, or contents of input. It shows us the complexity described in 0 to 1 of elements in a number set. Shannon entropy is used in Artificial intelligence, especially in decision trees. Shannon entropy has never been used for facial recognition system. However, in this paper, Shannon entropy will have a significant role in new facial recognition system.

## 2. THEORY

### 2.1. Fuzzy Theory

Fuzzy logic in Fuzzy theory is in contrast with traditional Crisp logic, which only gets true or false for the question, which means it is limited in applying to the real world. There are three types of shapes taken by Fuzzy number used in Fuzzy Theory.

1. Bell-curve shape
   Bell-curve shape's graph's shape is smooth. Its fuzzy number's degree of membership $\mu_A(x)$ is

   $$\mu_A(x) = [1 - \frac{(x-r)^2}{r^2}]e^{-\frac{(x-r)^2}{r^2}}$$

   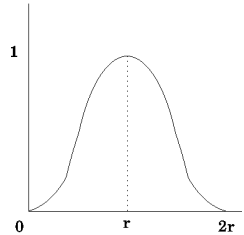

2. Triangle shape
   Triangle shape is a simplified version of Bell-curve shape, and is normally used. Its fuzzy number's degree of membership $\mu_{tr}(x)$ is

   $$\mu_{tr}(x) = \begin{cases} \frac{1}{(r-p)}(x-r)+1 & p < x \leq r \\ -\frac{1}{(q-r)}(x-r)+1 & r < x \leq q \end{cases}$$

   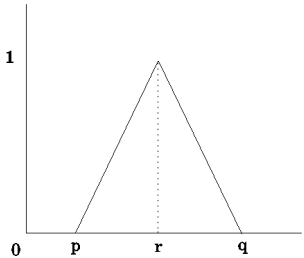

3. Trapezoid shape
   Trapezoid shape is combination of Triangle shape and Bell curve's shape and its graph's shape is simple. Its fuzzy number's degree of membership $\mu_{tz}(x)$ is

   $$\mu_{tz}(x) = \begin{cases} \frac{1}{(s-p)}(x-s)+1 & p < x \leq s \\ 1 & s < x \leq t \\ -\frac{1}{(q-t)}(x-t)+1 & t < x \leq q \end{cases}$$

   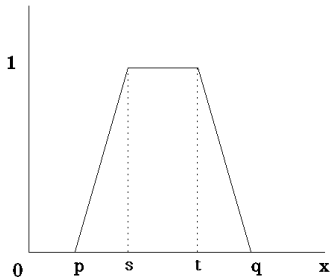

*The membership function shows the closeness of x to r in a number from 0 to 1.

## 2.2. Shannon's Entropy

Shannon's Entropy is a measure of uncertainty used in statistics, probability theory, computer science, and in statistical dynamics. It does a calculation with ratios, which means that even with different size of pictures, it wouldn't get bothered, which leads to higher accuracy on different sizes of images for facial recognition system. Shannon's entropy H(X) is

$$H(X) = -\sum_{i=1}^{n} p_i \log_2 p_i$$

$p_i$ is dividing $i^{th}$ number of set X by sum of all numbers in set X, which is probability of the $i^{th}$ number in set X.

## 3. METHOD

1. Calculate the length between two eyes, between end of nose and start of mouth, length between ears, length of mouth, length of eyebrow, and length from chin to middle of eyebrows, etc. All features are put to each set of features. (feature set:$X_1, X_2, X_3 \cdots$)

2. Draw face line until ears only for parts where is recognizable with computer. If size of input 1 and input 2 different, make the small one as large as the large input. And then, place lines in the same location in image. Subtract input 2 from input 1. Unless the area of result of subtraction is 0, $\alpha$=(leftover's area)/(input 1's area). If leftover is 0, Subtract input 1 from input 2. If the area of the result is not 0, $\alpha$=(leftover's area)/(input 2's area). If leftover area is 0 again, $\alpha$=1. [7]

3. Calculate each entropy for each feature sets to compare each features for each input. Since this is comparing features, this calculation will be crucial for the system.

$$H(X) = -\sum_{i=1}^{2} p_i \log_2 p_i \ (X = X_1, X_2, X_3 \cdots)$$

**Equation 1**

$$p_i = \frac{i^{th} \text{ element of X}}{\text{sum of all the elements in X}}$$

4. Calculate Bell-curve shape's fuzzy number's degree of membership $\mu_A(x)$ for each feature sets. Since there are n feature sets, the calculation will be held for n times.

$$\mu_A(x) = \left[1 - \frac{(x-r)^2}{r^2}\right] e^{-\frac{(x-r)^2}{r^2}}$$

(r=1,x=H(X))($X = X_1, X_2, X_3 \cdots X_n$)

**Equation 2**

5. Add up $\mu_A(x)$, and then divide the sum by n. Let's say the result of this is $\beta$. Than, result of this algorithm, $\delta$, which is similarity percentage of two inputs, will be like this:

$$\delta = 100 * (\beta * K^* + \alpha * (1 - K))$$

**Equation 3**

## 4. RESULT

Constant K is a number earned by the training. When gets two images with same person is inputted, the algorithm gets an instant variable $t_1, t_2$. Variable $t_1$ will be representing the number that outputs 0.95 as a result of method 7, and variable $t_2$ will be representing the number that outputs 1 as a result of method 7. The first value of variable $K_1$ will be variable $t_1$ of first input data, and the first value of variable $K_2$ will be variable $t_2$ of first input data. For other input data that are not the first, if $t_1$ is bigger than $K_1$ and smaller than $K_2$, $K_1$ will be $t_1$. If $t_2$ is smaller than $K_2$ and bigger than $K_1$, $K_2$ will be $t_2$. Else, $nK_1$ will be ($t_1$ +n* $K_1$)/(n+1), and $nK_2$ will be ($t_2$+n* $K_2$)/(n+1). Number n is the number of data sets given for training until previous data. The number K will be ($nK_1$+ $nK_2$)/2.

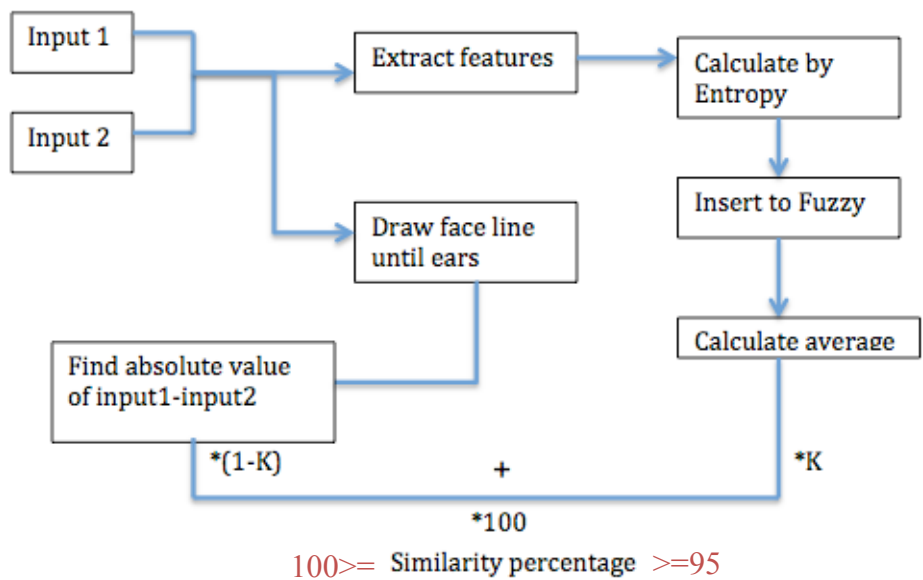

Figure <Summary of Method>

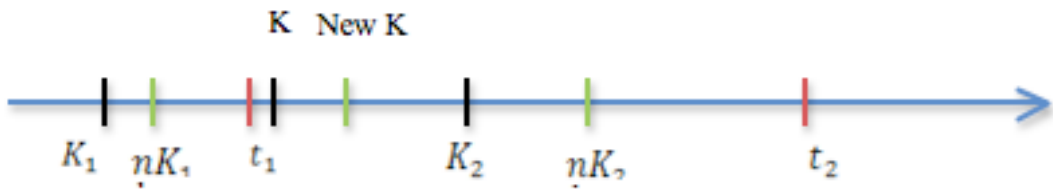

Figure <Range of numbers in Result>

## 5. CONCLUSION

This paper describes a new approach for facial recognition system by using entropy and fuzzy. This new method is much more simple than other methods described in [1][2][3][4][5][6]. In contrast with use of width and length of the picture with full-size face to compare the image [4], by using entropy, we now don't have to consider the size of the person's face on the picture unless it is recognizable to humans. Since Entropy calculates the ratio of whole and its one number, even if the size is changed. By considering the face shape only until ear, hair change wouldn't bother figuring out the person's identity. So, this system can be used for facial recognition security system, transforming the algorithm a bit for the real use. For example, for cell phone facial recognition system, each time of facial recognition, the image saved for recognition will be updated to the new one compared for unlock, so that the algorithm may react to the change of the person. The only problem with this algorithm is that it can't recognize the image that is not heading front. Future work will be directed on studying how to recognize 3D components in 2D picture, and apply that to this algorithm, increasing the accuracy of my new facial recognition system.

## REFERENCES

[1] Ara V. Nefian and Monson H. Hayes III (1998) 'Hidden Markov models for face recognition', *Proceedings, International Conference on Acoustics, Speech and Signal Processing,* pp. 2721-2724.

[2] D. Beymer, "Face recognition under varying pose," in Proceedings of 23rd Image Understanding Workshop, vol. 2, pp. 837-842, 1994.

[3] Haripriya Ganta and Pinky Tejwani.," Face Recognition using Eigenfaces". ECE 847 Project Report.. Fall 2004, Clemson University.

[4] John MacCornick (2013) *Nine Algorithms That Changed the Future: The Ingenious Ideas That Drive Today's Computers*, 41 William Street, Princeton, New Jersey 08540: Princeton University Press.

[5] Keun-Chang Kwak, Hyoun-Joo Go, Myung-Geun Chun (2004) 'Face recognition using Fisherface Method with Fuzzy Membership Degree. ', *Journal of KIISE : Software and Applications,* vol.31(6th), pp. 784-791.

[6] Won-Suk Chang, Chang-Hyeon Noh, Jong-Sik Lee (2009) 'Fuzzy-based Segment-Boost Method for Effective Face Recognition ', *JOURNAL OF THE KOREA SOCIETY FOR SIMULATION,* vol.18(1st), pp. 17-25.

[7] Yeunghak Lee (2008) '3D Face Recognition in the Multiple-Contour Line Area Using Fuzzy Integral', *JOURNAL OF KOREA MULTIMEDIA SOCIETY,* vol.11(4th), pp. 423-433.

**Jaejun Lee**

was born in Seoul, Korea, in 1998. He is currently a student in science major of Hankuk Foreign Studies ,Korea. He is mostly interested in computer science and has been studying pattern analysis and computer programming.

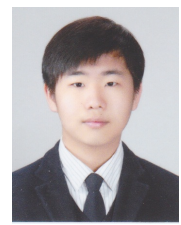

**Taeseon Yoon**

was born in Seoul, Korea, in 1972. He was Ph.D. Candidate degree in Computer education from the Korea University, Seoul, Korea, in 2003.
From 1998 to 2003, he was with EJB analyst and SCJP. From 2003 to 2004, he joined the Department of Computer Education, University of Korea, as a Lecturer and Ansan University, as a Adjunct professor. Since December 2004, he has been with the Hankuk Academy of Foreign Studies, where he was a Computer Science and Statistics Teacher. He was the recipient of the Best Teacher Award of the Science Conference, Gyeonggi-do, Korea, 2013

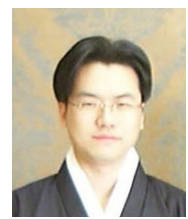